\documentclass[3p,times,preprint,11.5pt]{elsarticle}

\usepackage{amssymb}
\usepackage{amsmath}
\usepackage{multirow, multicol}
\usepackage{hyperref}
\usepackage{tabularray}
\usepackage{graphicx}
\usepackage{textcomp}
\usepackage{algorithm}
\usepackage{algorithmic}
\usepackage{amsmath}
\usepackage{booktabs}
\usepackage{xcolor}
\usepackage{color}
\usepackage{colortbl, booktabs}
\hypersetup{
    colorlinks=true, 
    urlcolor=magenta,   
}

\begin{document}

\begin{frontmatter}



\title{Exploiting Point-Language Models with
Dual-Prompts for 3D Anomaly Detection}


\author[label1,label3]{Jiaxiang Wang\corref{cor1}}
\ead{wangjiaxiang@xmu.edu.cn}
\author[label1,label2]{Haote Xu\corref{cor1}}
\ead{hotxu@stu.xmu.edu.cn}

\author[label1,label2]{Xiaolu Chen}
\author[label4]{Andreas Jakobsson}
\author[label1,label2]{Haodi Xu}
\author[label1,label2]{Yue Huang}
\author[label1,label2]{Xinghao Ding}
\author[label1,label2]{Xiaotng Tu\corref{cor1}\footnote{Corresponding author.}}
\ead{xttu@xmu.edu.cn}

\cortext[cor1]{All the authors contribute equally to this work.}


\affiliation[label1]{organization={Key Laboratory of Multimedia Trusted Perception and Efficient Computing, Ministry of Education of China},
            addressline={Xiamen University},
            city={Xiamen},
            postcode={361005},
            state={Fujian},
            country={China}}

\affiliation[label2]{organization={School of Informatics},
            addressline={Xiamen University},
            city={Xiamen},
            postcode={361005},
            state={Fujian},
            country={China}}

\affiliation[label3]{organization={Institute of Artificial Intelligence},
            addressline={Xiamen University},
            city={Xiamen},
            postcode={361005},
            state={Fujian},
            country={China}}
\affiliation[label4]{organization={Mathematical Statistics},
            addressline={Lund University},
            country={Sweden}}
\begin{abstract}
Anomaly detection (AD) in 3D point clouds is crucial in a wide range of industrial applications, especially in various forms of precision manufacturing. Considering the industrial demand for reliable 3D AD, several methods have been developed. However, most of these approaches typically require training separate models for each category, which is memory-intensive and lacks flexibility. In this paper, we propose a novel \underline{P}oint-\underline{L}anguage model with dual-prompts for 3D \underline{AN}omaly d\underline{E}tection (PLANE). The approach leverages multi-modal prompts to extend the strong generalization capabilities of pre-trained Point-Language Models (PLMs) to the domain of 3D point cloud AD, achieving impressive detection performance across multiple categories using a single model. Specifically, we propose a dual-prompt learning method, incorporating both text and point cloud prompts. The method utilizes a dynamic prompt creator module (DPCM) to produce sample-specific dynamic prompts, which are then integrated with class-specific static prompts for each modality, effectively driving the PLMs. Additionally, based on the characteristics of point cloud data, we propose a pseudo 3D anomaly generation method (Ano3D) to improve the model's detection capabilities in unsupervised setting. Experimental results demonstrate that the proposed method, which is under the multi-class-one-model paradigm, achieves a +8.7\%/+17\% gain on anomaly detection and localization performance as compared to the state-of-the-art one-class-one-model methods for the Anomaly-ShapeNet dataset, and obtains +4.3\%/+4.1\% gain for the Real3D-AD dataset. Code will be available upon publication.
\end{abstract}




\end{frontmatter}



\section{Introduction}
\label{sec:introduction}
The area of 2D Anomaly Detection (AD) has attracted notable attention during the past decades\cite{RD++,jia2024unsupervisedti1,shang2023defect,bae2023pni}. With the increasing importance of 3D point clouds in autonomous driving and robotics, research on 3D point cloud AD has also garnered growing attention. \cite{liu2024real3d,IMR,li2024daup}. This phenomenon can be explained by the fact that, compared to 2D images, point cloud data encapsulates richer structural and spatial information, enabling more detailed and precise detection.

\begin{figure}[t]
\centering
\includegraphics[width=0.6\columnwidth]{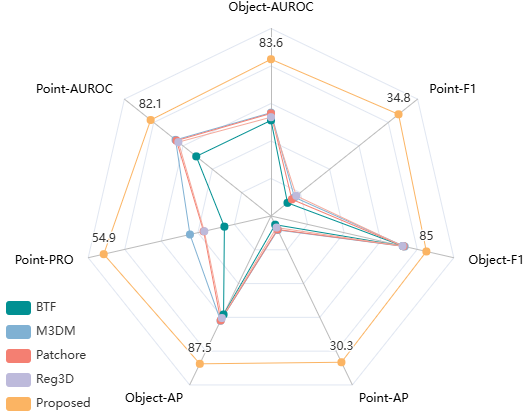}
\caption{Radar chart of comparisons in terms of evaluation metrics between baselines and the proposed methods on the Anomaly-ShapeNet dataset.}
\label{fig:1}
\end{figure}

Current methods for 3D AD predominantly employ various forms of unsupervised approaches \cite{r3dad,IMR,wang2023m3dm}, wherein only normal samples are utilized during training to tackle the challenges posed by the variability of anomalies and the difficulty in collecting anomalous samples. These methods are mainly divided into two categories: reconstruction and embedding-based approaches, both of which have evolved from recent 2D AD techniques.
Reconstruction methods are based on the premise that models trained on normal samples cannot effectively reconstruct anomalous ones\cite{IMR,r3dad}. During the inference phase, anomalies are then detected by comparing the original samples with their reconstructions.
However, due to higher acquisition costs, 3D point cloud data is typically less abundant than 2D data. This scarcity of training data often leads to overfitting, which results in poor performance for these models.
Embedding-based methods utilize pre-trained networks to construct memory banks of normal features for AD\cite{roth2022patchcore,yang2023memseg,lee2022cfa}. During the inference phase, the anomaly score is then determined by comparing the similarity between the features of the test samples and the features stored in the memory bank for normal data. Nevertheless, this memory bank grows incrementally with the number of samples, and the requirement to compare each feature during inference significantly increases processing time, rendering these approaches impractical. Furthermore, current methods generally adhere to a one-class-one-model paradigm, which not only increases storage costs but also limits flexibility. To alleviate these shortcomings, we propose a novel 3D point cloud AD approach that is both accurate and efficient, while also capable of handling multi-class detection within a single model.

 Vision-language models (VLMs) \cite{CLIP,li2022blip,della} pre-trained on large-scale vision-language datasets can effectively transfer knowledge to various downstream tasks, achieving impressive results, without the need for full-parameter fine-tuning. Similarly, in the 3D point cloud domain, specialized Point Language Models (PLMs) \cite{zhu2023pointclip2,xue2024ulip} have been developed, suggesting that relying on the robust representation capabilities of PLMs is a viable approach for achieving multi-class 3D point cloud AD within a single model, particularly in scenarios where the training data is limited.

Inspired by existing 2D AD methods utilizing VLMs \cite{jeong2023winclip,li2024promptad,cao2025adaclip}, we apply PLMs (e.g., ULIP2 \cite{xue2024ulip}) to perform AD by measuring the similarity between multi-level point cloud and text features. The effectiveness of this approach relies on the precise re-alignment of features across different modalities during training. However, current 2D AD VLMs-based methods are suboptimal because they adjust features in a single modality (language or vision) for unidirectional alignment, which prevents the dynamic alignment of both representation spaces on the AD task. Hence, we propose the \textbf{P}oint-\textbf{L}anguage model with dual-prompts for 3D \textbf{AN}omaly d\textbf{E}tection (PLANE) approach. Specifically, recognizing the minimal intra-class sample variations characteristic of AD, we construct learnable class-specific static prompts for each modality. In addition, we employ the dynamic prompt generation module to create sample-specific dynamic prompts, which are then integrated with the static prompts for each modality. This approach promotes interaction between modality-specific prompts and further enhances the alignment of multimodal features. Additionally, existing methods demonstrate that representations learned by detecting irregularity introduced by synthetic anomalies generalize well to detecting real defects\cite{xu2024afscti3,zavrtanik2021draem,liu2025simple}.  In this way, we propose a pseudo 3D anomaly generation method for 3D point cloud data, called Ano3D, to enhance the AD models' robustness and generalization capabilities. In summary, the contributions of our work include:
\begin{itemize}
\item To the best of our knowledge, we are the first to introduce PLMs for the task of 3D point cloud AD, allowing us to develop a multi-class-one-model architecture that is both accurate and efficient.
\item We propose a dual-prompt approach that integrates point cloud and text prompts to drive the PLMs, leveraging a combination of dynamic and static prompts to achieve high re-alignment of multimodal features in 3D point cloud AD tasks.
\item We propose a method named Ano3D for generating synthetic 3D point cloud anomaly samples, and demonstrate through experiments that it generalizes well to real-world anomalies.
\item Experiments on the Anomaly-ShapeNet and Real3D-AD datasets show that PLANE outperforms existing state-of-the-art methods under the more challenging multi-class-one-model paradigm, achieving superior results across multiple evaluation metrics, as illustrated in Fig. \ref{fig:1}.
\end{itemize}
The remainder of the article is organized as follows: the related work on 2D and 3D AD is reviewed in Section II. Then we introduce our proposed method PLANE in detail in Section III. Section IV provides the experimental results and their analysis, and Section V concludes the article.

\section{Related Work}
This study categorizes existing industrial AD tasks into two primary categories: 1) 2D and 2) 3D AD, which are introduced as follows.
\subsection{2D Anomaly Detection} 
2D AD methods can be mainly categorized into reconstruction and embedding-based approaches. Reconstruction methods assume that models are unable to reconstruct anomalous images because of training only on normal images\cite{zavrtanik2021draem,zhang2024realnet,xu2024afscti3}. DRAEM \cite{zavrtanik2021draem} trains a U-Net network for reconstruction using artificially generated pseudo-anomalies and employs a discriminator network to distinguish anomalous regions. Embedding-based methods can be further divided into memory bank approaches and vision-language models approaches\cite{RD,roth2022patchcore}. Patchcore \cite{roth2022patchcore} uses a pre-trained feature extractor to build a greedy coreset of representative normal features. VLMs methods leverage the powerful generalization capabilities of pre-trained models\cite{jeong2023winclip,li2024promptad}. WinClip \cite{jeong2023winclip} utilizes the pre-trained CLIP model to effectively extract and aggregate multi-scale text-aligned features. PromptAD \cite{li2024promptad} addresses the limitations of manually setting text prompts by introducing prompt learning, thereby constructing suitable text prompts to guide the model in performing AD more effectively.
\subsection{3D Anomaly Detection} 
3D AD is a more challenging task than 2D AD due to its typically unordered nature and high level of sparsity. Derived from 2D AD methods, AD in 3D can also be divided into reconstruction and embedding-based. IMRNet, which is inspired by the masking strategy, trains a point cloud reconstruction network in a self-supervised manner\cite{IMR}.  R3D-AD uses a diffusion model to transform the point cloud reconstruction task into a conditional generation task, detecting anomalies through a trained reconstruction network\cite{r3dad}. M3DM detects and localizes anomalies through unsupervised feature fusion, combining multimodal information with multiple memory banks\cite{wang2023m3dm}. CPMF renders 3D point cloud data into 2D images from multiple viewpoints and builds a memory bank of complementary multimodal features\cite{cao2024cpmf}. Reg3DAD constructs a feature memory bank of normal point cloud samples using a pre-trained 3D feature extractor\cite{liu2024real3d}. CFM4IAD achieves mutual mapping between 2D and 3D features through fully connected layers, facilitating collaborative detection\cite{CFM4IAD}. The CLIP3D-AD method \cite{zuo2024clip3dad} leverages point cloud projection to eliminate the modality gap with the pre-trained CLIP \cite{CLIP} model, enhancing vision and language correlation through the fusion of multi-view image features.

\section{Methodology}
As illustrated in Fig. 2, the proposed PLANE method primarily consists of a pretrained PLM, along with learnable text and point cloud prompts. These dual multimodal prompts are composed of both static and dynamic components. Leveraging this prompt configuration, the pre-trained PLM extracts point cloud and text features, which are then projected onto a joint embedding space via a feature adaptation layer. Finally, anomalous regions are detected by measuring the similarity between multi-level point cloud features and text features. Additionally, to address the issue of overfitting due to insufficient data, we propose a pseudo 3D anomaly generation (Ano3D) method for data augmentation. This approach expands the training dataset, enhancing the model's generalization ability. The details are explained in the following sections.

\begin{figure}[t]
\centering
\includegraphics[width=0.85\textwidth]{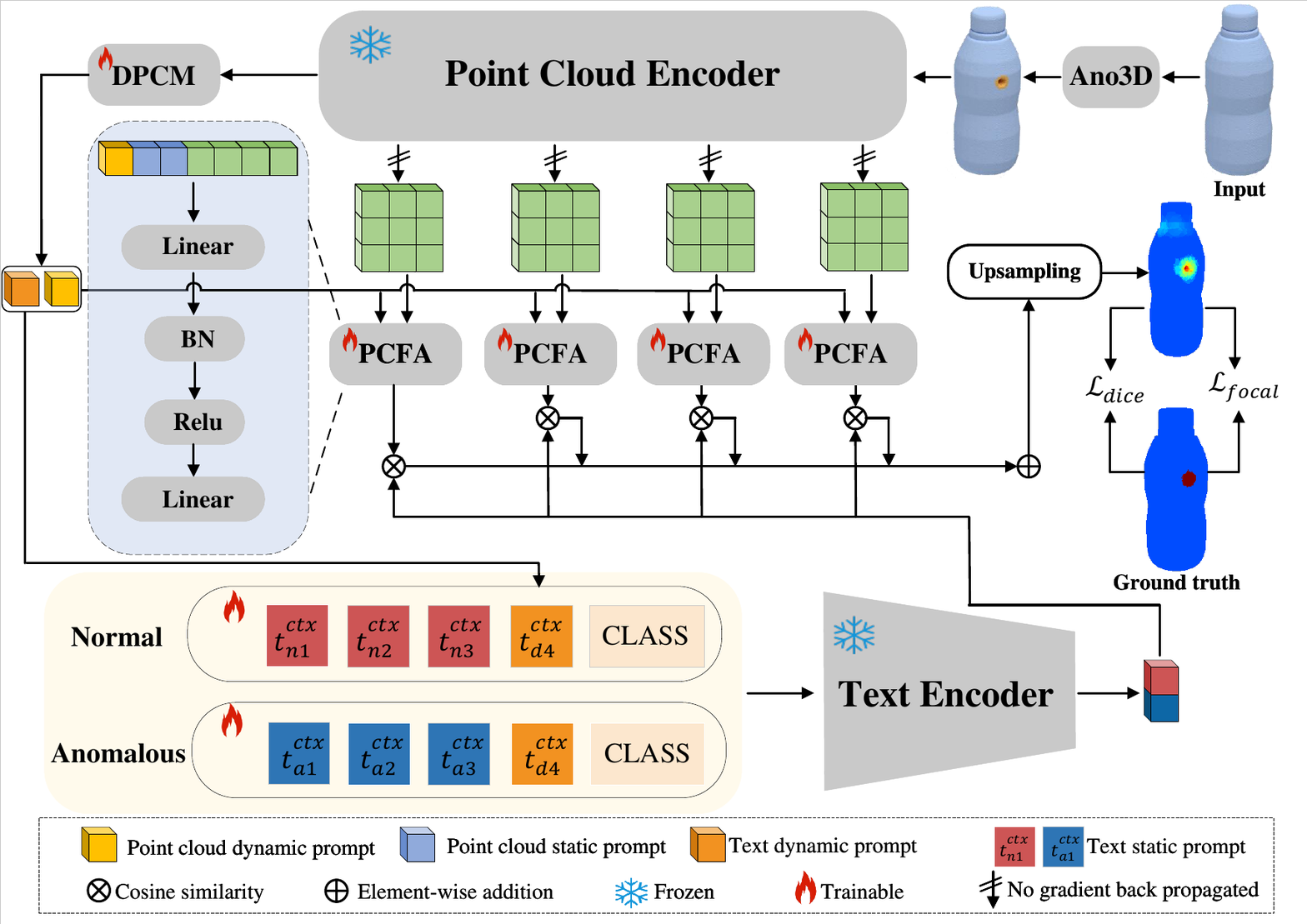}
\label{fig:2}
\caption{Overview of the PLANE pipeline. During the training phase, samples are first processed through the Ano3D module to generate pseudo-anomalous point cloud samples, which are then input into a pre-trained point cloud encoder. Next, the output of the encoder is fed into the DPCM module to generate corresponding dynamic text and point cloud prompts. Subsequently, the intermediate features from the point cloud encoder, along with the static and dynamic point cloud prompts, are fed into the PCFA module, where they are projected into a joint embedding space with text features. Meanwhile, the dynamic text prompts generated by the DPCM module are combined with static text prompts and processed through the pre-trained text encoder to produce the final text features. Finally, the model calculates the similarity between multi-hierarchical point cloud features and text features, aggregates the results, and compares them with ground truth to calculate the loss, thereby training the model.}
\end{figure}

\subsection{Dual-Prompt Learning}


Earlier efforts have demonstrated the effectiveness of using integrated handcrafted prompts for AD. However, the capacity of these prompts is inherently limited, as natural language cannot capture the full scope of detailed vision information. This limitation may lead to a misalignment between fine-grained point cloud features and text features, especially in tasks that require precise localization of anomalous regions. To mitigate this issue, some methods have introduced learnable text prompts in 2D AD tasks \cite{li2024promptad,zhou2023anomalyclip}, but these approaches allow only for a unidirectional adjustment of the text representation space. In response, this paper introduces dual prompts that bidirectionally adjust the representation spaces of both modalities within the PLM, improving their suitability for 3D AD tasks. The following section provides a detailed discussion of the prompts for each modality.

In contrast to previous designs of learnable text prompts, the proposed approach incorporates both static and dynamic components in the normal prompt $S^{n}$ and in the anomalous prompt  $S^{a}$:
\begin{center}
    \begin{equation}\label{eq1}
    \centering
S^{n} = [P^{n}_{1}][P^{n}_{2}][P^{n}_{3}][P^{n}_{N}][P_{dyn}][obj],
    \end{equation} 
\end{center}
   
\begin{equation}\label{eq2}
S^{a} = [P^{a}_{1}][P^{a}_{2}][P^{a}_{3}][P^{a}_{N}][P_{dyn}][obj],
\end{equation}  
where $N$ denotes the length of learnable static prompts, $P_{dyn}$ denotes the dynamic prompts, and the symbol $[obj]$ denotes the category’s name. 
It is noteworthy that, in addition to text prompts, the prompts for the point cloud modality also include both dynamic and static components. This design is based on the observation that variations among samples within the same class are minimal in AD tasks. As a result, we implement the class-specific static prompts for point cloud modality, allowing objects of the same class to share a consistent set of prompt components. To further refine the alignment between the two modalities, sample-specific dynamic prompts are introduced, generated by the dynamic prompt creator module (DPCM).
It is worth noting that the final layer features of the point cloud encoder encapsulate critical global information. Therefore, the DPCM maps these features through a Multi-Layer Perceptron (MLP) and slices them to generate both point cloud dynamic prompts and text dynamic prompts. Given an input point cloud represented as $P_n$, the DPCM process can be formalized as:
\begin{align}
f_{pc} &= PointEncoder(P_n), \label{eq1} \\
f_{dyn} &= MLP(f_{pc}). \label{eq2}
\end{align}
where $f_{dyn}$ denotes dynamic prompts, we then split $f_{dyn}$ into two parts, $f_{dyn} = \{f_{d}^{t}, f_{d}^{p}\}$, and $f_{d}^{t}$ and $f_{d}^{p}$ denote dynamic text prompts and dynamic point cloud prompts, respectively. The dynamic prompts generated from point cloud samples are combined with static prompts from their respective modalities, enhancing the correlation between the two modalities.

The pre-trained point cloud encoder is primarily designed for classification, and its intermediate layer features are not aligned with text data during pre-training. This misalignment limits its capacity for fine-grained detection in AD tasks. To address this, the Point Cloud Feature Adaptation (PCFA) module is introduced to project the intermediate point cloud features into a joint space shared with text features. Within PCFA, the intermediate layer features of the point cloud, along with class-specific static point cloud prompts and sample-specific dynamic point prompts, serve as inputs. A feedforward network (FFN) then integrates the point cloud features $f_{ori}$ and point cloud prompts to achieve alignment with the text features. This process may be expressed as:
\begin{equation}\label{eq3}
F^{P}_{i} = FFN([P^{vis}_{sta}, P^{vis}_{dyn}, f_{ori}]),
\end{equation}   
where $P^{vis}_{sta}$ represents the class-specific static prompt, $P^{vis}_{dyn}$ the sample-specific dynamic prompt, and $[$· , ·$]$ the concatenation operation. By utilizing PCFA and the dual-prompt method to bidirectionally realign point cloud and text features, the AD result at the $i$-th hierarchy can be generated as:

{\small
\begin{equation}\label{eq2}
    \mathbf{S}_{i}=\phi\left(\frac{\exp \left(\cos \left(\mathbf{F}_{i}^{P}, \mathbf{F}_{A}^{T}\right)\right)}{\exp \left(\cos \left(\mathbf{F}_{i}^{P}, \mathbf{F}_{N}^{T}\right)\right)+\exp \left(\cos \left(\mathbf{F}_{i}^{P}, \mathbf{F}_{A}^{T}\right)\right)}\right),
\end{equation}}where $\phi$ denotes the reshape and interpolate function, $\mathbf{F}_{i}^{P}$ represents point cloud features, whereas $\mathbf{F}_{N}^{T}$ and $\mathbf{F}_{A}^{T}$ indicate normal text features and anomalous text features, respectively. 

\subsection{Pseudo 3D Anomaly Generation}


In this section, we introduce the Ano3D method for Pseudo 3D anomaly generation, aimed at augmenting the training dataset. This method simulates three common types of anomalies (e.g., hole, bulge, and concavity) based on the characteristics of 3D point clouds. 

Specifically, we first randomly initialize the rotation angles ($\theta_{x}$, $\theta_{y}$, $\theta_{z}$) to obtain the rotation matrix. By multiplying the original point cloud sample with this rotation matrix, the rotated point cloud sample is generated. For protrusions and indentations, a center point is randomly selected, followed by the selection of $M$ neighboring points using nearest-neighbor sampling.
Since the orientations of actual protrusions and indentations are typically perpendicular to the plane, the normal vector perpendicular to the selected surface points is calculated. A normal distribution is then sampled to determine the displacement distances. The surrounding $M$ points are translated by a certain distance along the normal vector, while the rest of the points remain unchanged. If the normal vector points inward, a concavity defect is simulated; if it points outward, a bulge defect is created. For simulating hole defects, a point is randomly selected in the point cloud, and the nearest $X$ points around it are identified. These $X+1$ points are then removed to create a pseudo-anomalous hole. This method of constructing pseudo-anomaly samples allows for the generation of diverse and realistic anomaly examples. The pseudo-code of Ano3D is summarized in Algorithm \ref{algorithm:1}.

\begin{figure}[t]
\centering
\includegraphics[width=0.8\columnwidth]{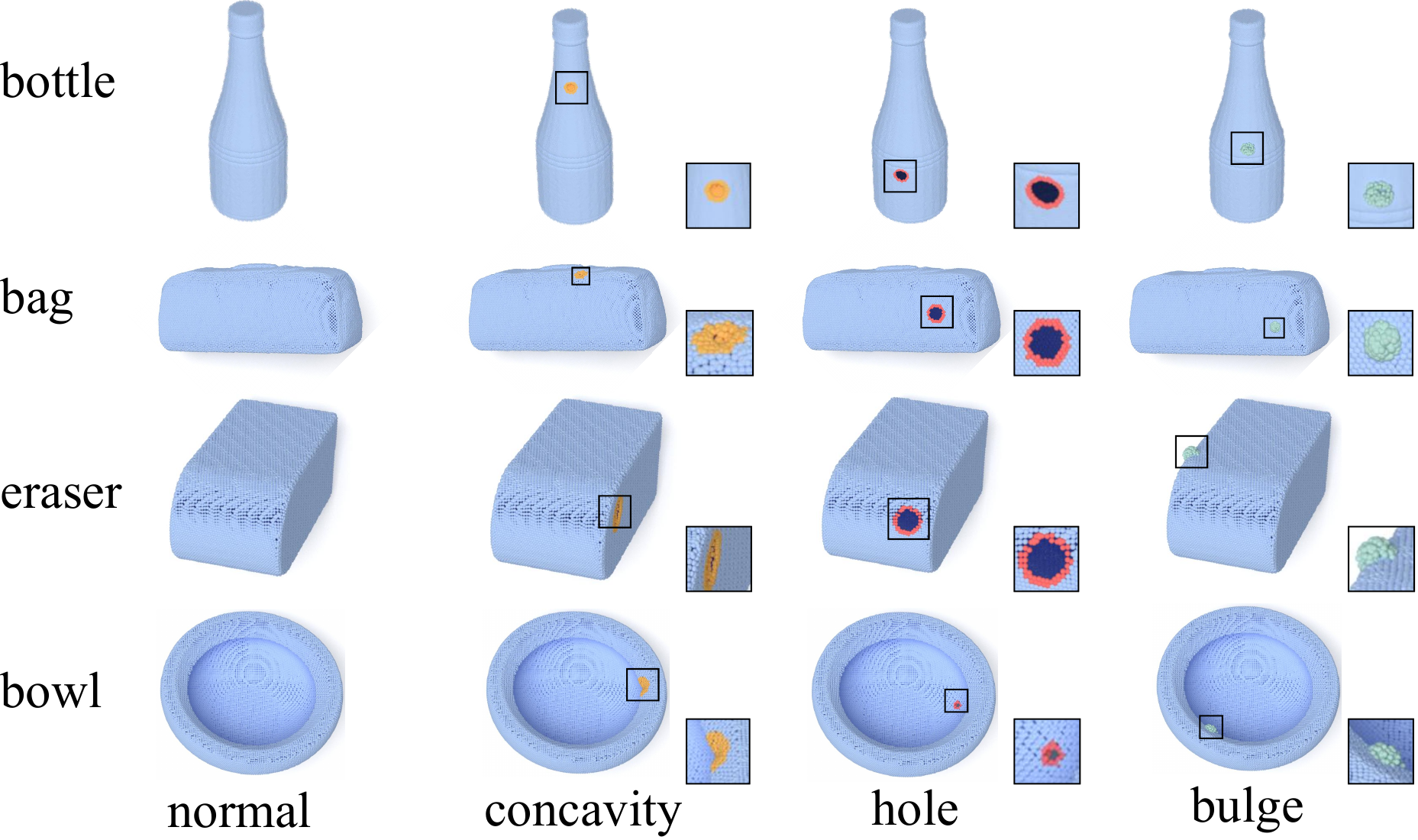}
\caption{Synthetic anomalies created with Ano3D}
\label{fig:3}
\end{figure}

Examples of pseudo-anomaly samples constructed by the Ano3D method are shown in Fig. \ref{fig:3}, where the first column represents normal point cloud samples, and the last three columns depict pseudo-anomaly point cloud samples generated by Ano3D. As can be seen, Ano3D simulates various defects, including concavities, holes, and bulges, which are anomalies specifically tailored to the characteristics of 3D point cloud data. Incorporating these pseudo-anomalous samples during training alleviates the scarcity of anomalous point cloud samples in the training set and enhances the model's generalization ability.

\begin{algorithm}[tb]
\caption{Ano3D}
\label{alg:algorithm}
\begin{algorithmic} 
\FOR{epoch in 1,2,...,n}
    \FOR{$P_{i}$ in normal training-set S}
\STATE Random init rotation angles $\theta_{x}$, $\theta_{y}$, $\theta{z}$
\STATE $R$ = CreateRotationMatrix($\theta_{x}$, $\theta_{y}$, $\theta{z}$)
\STATE $P^{r}_{i}$ = $P_{i} \cdot R$

\STATE $j$ = $randint(0, P^{r}_{i}.shape[0])$ \textcolor{gray}{// select central point}
\STATE $P^{n}_{i}$, $index$ = $NN(P^{r}_{i}[j], P^{r}_{i}, M)$ \textcolor{gray}{//  $M$ nearest  points}
\STATE $\vec{n}$ = $GetNormalVector(P^{n}_{i})$
\STATE $\mathcal{T}$ $\thicksim$  $\mathcal{N}(\mu,\sigma^2)$ \textcolor{gray}{// sampling $\mathcal{T}$ from $\mathcal{N}(\mu,\sigma^2)$.}
\STATE $\epsilon$ = $np.zeros(P^{r}_{i}.shape)$
\STATE  $\epsilon[index]$ = $\vec{n} \odot \mathcal{T}$
\STATE $P^{r}_{i} = P^{r}_{i}  +  \epsilon$ \textcolor{gray}{// augmented point cloud}
\STATE $P^{r}_{i}$ as model input and executive training process
\ENDFOR
\ENDFOR
\end{algorithmic}
\label{algorithm:1}
\end{algorithm}

\subsection{Loss Function}

The proposed PLANE method is an end-to-end trainable model. In the approach, the text encoder and point encoder models are frozen. To train the projection layer and learnable prompts, two loss functions are employed using focal loss and dice loss. Focal loss \cite{lin2017focal} addresses the issue of class imbalance, which is particularly relevant in 3D AD where the majority of points are normal and only a small portion are anomalous. Therefore, focal loss is utilized to mitigate this imbalance. Dice loss \cite{dice}, derived from the dice coefficient and commonly used in image segmentation tasks, can alleviate the class imbalance between normal and anomalous points. The total loss function for PLANE is then calculated as:

\begin{equation}\label{eq3}
L_{total} = Focal(\sum_{i=1}^{L}\mathbf{S}_{i},G) +  Dice(\sum_{i=1}^{L}\mathbf{S}_{i},G),
\end{equation}   
where $S_{i}$ denotes the ${i}$-th intermediate anomaly map and $G$ the ground truth corresponding to the input point cloud.

During the testing phase, anomaly scores from multiple layers are accumulated and upsampled to match the scale of the input point cloud, producing an anomaly score map for the sample. The maximum value within this map serves as the anomaly score for the entire sample.

\section{Experiment}
\subsection{Datasets and Evaluation Metrics}
\textbf{Anomaly-ShapeNet Dataset.} The Anomaly-ShapeNet \cite{IMR} dataset is a synthesized 3D AD dataset based on ShapeNetCorev2 \cite{chang2015shapenet}, which includes 40 categories with over 1,600 samples. Similar to the few-shot setup in 2D AD, each category contains only 4 normal training samples, and the test set includes both normal and anomalous samples, with each sample consisting of between 8,000 and 30,000 points. It encompasses six different types of anomalies, which significantly increases the diversity of anomaly types.

\textbf{Real3D-AD Dataset.} The Real3D-AD \cite{liu2024real3d} dataset is a real 3D AD dataset with 1,254 samples among 12 categories. Each category contains only 4 training samples. The training set provides a complete 360-degree scan of the objects without any blind spots. The anomaly samples are obtained through one-sided scanning, which aligns with the actual scenarios in an industrial inspection while increasing the challenge  of AD.

\textbf{Evaluation Metrics.} We measure the detection of objects-level anomalies and the location of points-level anomalies using metrics such as the area under the receiver operating characteristic curve (AUROC), the average precision (AP) and the F1 score max (F1). In addition, for anomaly localization, we report the Area Under the Per-Region Overlap Curve (AU-PRO).

\begin{table}
  \centering
  \scriptsize
  \renewcommand\arraystretch{1.0}
  \resizebox{\linewidth}{!}{
  \begin{tabular}{@{}cl|m{0.6cm}<{\centering}m{0.6cm}<{\centering}m{0.6cm}<{\centering}m{0.6cm}<{\centering}m{0.6cm}<{\centering}m{0.6cm}<{\centering}m{0.6cm}<{\centering}m{0.65cm}<{\centering}m{0.65cm}<{\centering}m{0.6cm}<{\centering}m{0.6cm}<{\centering}m{0.6cm}<{\centering}m{0.6cm}<{\centering} m{0.6cm}<{\centering}}
    \toprule
    & Method  & ashtray0 & bag0 & bottle0 & bottle1 & bottle3 & bowl0 & bowl1 & bowl2 & bowl3 & bowl4 & bowl5 & bucket0 & bucket1 &cap0\\
    \midrule 
     &BTF (CVPR2023) & 0.529       & 0.557                          & 0.533          & 0.512        & 0.492       & 0.493       & 0.456       & 0.570        & 0.419       & 0.511       & 0.505       & 0.394       & 0.489 & 0.456 \\
     &M3DM$_{MAE}$ (CVPR2023)  & 0.600         & 0.652                          & 0.662          & 0.561        & 0.625       & 0.567       & 0.581       & 0.441       & 0.444       & 0.615       & 0.470        & 0.689       & 0.451 & 0.567 \\
     &M3DM$_{BERT}$ (CVPR2023)   & 0.571       & 0.605                          & 0.600            & 0.600          & 0.717       & 0.563       & 0.548       & 0.478       & 0.389       & 0.556       & 0.554       & 0.527       & 0.489 & 0.467  \\
     &Patchcore$_{MAE}$ (CVPR2022) & 0.610        & 0.671                          & \underline{0.733}          & 0.582        & 0.625       & 0.570        & 0.470        & 0.363       & 0.467       & 0.730        & 0.681       & 0.571       & 0.368 & 0.456 \\
     &Patchcore$_{BERT}$ (CVPR2022) & 0.695       & 0.652                          & 0.705          & 0.551        & 0.717       & 0.604       & 0.511       & 0.511       & 0.459       & 0.567       & 0.607       & \underline{0.695}       & 0.384 & 0.500 \\
     &Reg3DAD (NIPS2023)  & 0.533       & 0.652                          & 0.695          & 0.639        & 0.584       & 0.552       & 0.504       & 0.393       & 0.433       & 0.563       & 0.551       & 0.476       & 0.470 & 0.407\\
     &IMR (CVPR2024)  & 0.671       & 0.660                          & 0.552          & 0.700        & 0.640       & 0.681       & 0.702       & 0.685       & 0.599       & 0.676       & \underline{0.710}       & 0.580       & \underline{0.771} & 0.737\\
     &R3DAD (ECCV2024)  & \underline{0.833}       & \underline{0.720}                          & \underline{0.733}          & \underline{0.737}        & \underline{0.781}       & \underline{0.819}       & \underline{0.778}       & \underline{0.741}       & \textbf{0.767}       & \underline{0.744}       & 0.656       & 0.683       & 0.756 & \underline{0.822}\\
     &\textbf{PLANE} & \textbf{0.905}      & \textbf{0.914}                          & \textbf{0.843}          & \textbf{0.814}        & \textbf{0.994} & \textbf{0.963} & \textbf{0.907} & \textbf{0.956} & \underline{0.706} & \textbf{0.893} & \textbf{0.800}   & \textbf{0.981} & \textbf{0.968}          & \textbf{0.944} \\
     \midrule
    \midrule
    & Method & cap3        & cap4        & cap5        & cup0        & cup1        & eraser0     & headset0    & headset1    & helmet0     & helmet1     & helmet2     & helmet3     & jar0        & phone0\\
    \midrule 
     &BTF (CVPR2023)  & 0.491      & 0.625                          & \underline{0.691}          & 0.486        & 0.543 &0.586 & 0.489 & 0.481 & 0.580  & \underline{0.586} & 0.42  & 0.615 & 0.362 & 0.486 \\
     &M3DM$_{MAE}$ (CVPR2023)  &0.558      & \underline{0.684}                          & 0.537          & 0.581        & 0.500   & 0.667 & 0.502 & 0.486 & 0.464 & 0.476 & 0.464 & 0.506 & 0.686 & 0.733\\
     &M3DM$_{BERT}$ (CVPR2023)   & 0.533      & 0.632                          & 0.498          & 0.652        & 0.576 & 0.657 & 0.471 & 0.662 & 0.458 & 0.514 & 0.501 & 0.536 & 0.614 & \underline{0.781} \\
     &Patchcore$_{MAE}$ (CVPR2022) & 0.418      & 0.632                          & 0.456          & 0.614        & 0.567 & 0.567 & 0.387 & 0.576 & 0.533 & 0.476 & 0.554 & 0.503 & 0.662 & 0.686 \\
     &Patchcore$_{BERT}$ (CVPR2022) & 0.632      & 0.628                          & 0.523          & 0.567        & 0.600   & 0.638 & 0.520  & 0.514 & 0.586 & 0.467 & 0.551 & 0.518 & 0.671 & 0.738 \\
     &Reg3DAD (NIPS2023)  & 0.607      & 0.474                          & 0.519          & 0.571        & 0.490  & 0.576 & 0.378 & 0.600   & 0.623 & 0.410  & 0.513 & 0.533 & 0.614 & 0.600 \\
     &IMR (CVPR2024)  &\underline{0.775}      & 0.652                          & 0.652          & 0.643        & \textbf{0.757} & 0.548 & 0.720 & 0.676 & 0.597 & 0.600 & 0.641 & 0.573 & 0.780 & 0.755\\
     &R3DAD (ECCV2024)  & 0.730      & 0.681                          & 0.670          & \underline{0.776}        & \textbf{0.757} & \underline{0.890} & \underline{0.738} & \textbf{0.795} & \textbf{0.757} & \textbf{0.633} & \underline{0.707} & \underline{0.720} & \underline{0.838} & 0.762\\
     &\textbf{PLANE} & \textbf{0.954}      & \textbf{0.730}                           & \textbf{0.870}           & \textbf{0.805}        & \underline{0.705} & \textbf{1.000}     & \textbf{0.782} & \underline{0.776} & \underline{0.704} & 0.543 & \textbf{1.000}     & \textbf{0.721} & \textbf{1.000}              & \textbf{1.000}     \\
    \midrule
    \midrule
    & Method & shelf0      & tap0        & tap1        & vase0       & vase1       & vase2       & vase3       & vase4       & vase5       & vase7       & vase8       & vase9 & \multicolumn{2}{|c}{\textbf{Mean}} \\
    \midrule 
     &BTF (CVPR2023)  & 0.557      & 0.430                           & 0.404          & 0.500          & 0.476 & 0.600   & 0.573 & \underline{0.658} & 0.467 & 0.690  & 0.352 & 0.355&\multicolumn{2}{|c}{0.510} \\
     &M3DM$_{MAE}$ (CVPR2023)  & 0.464      & 0.500                            & 0.359          & 0.604        & 0.605 & 0.510  & 0.555 & 0.515 & 0.595 & 0.514 & 0.603 & 0.436&\multicolumn{2}{|c}{0.551} \\
     &M3DM$_{BERT}$ (CVPR2023)   & 0.554      & 0.536                          & 0.385          & 0.637        & 0.567 & 0.590  & 0.627 & 0.518 & 0.548 & 0.510  & 0.603 & 0.382&\multicolumn{2}{|c}{0.555} \\
     &Patchcore$_{MAE}$ (CVPR2022) & 0.426      & 0.556                          & 0.393          & 0.638        & 0.595 & 0.595 & 0.564 & 0.503 & 0.462 & 0.576 & 0.539 & 0.515&\multicolumn{2}{|c}{0.547} \\
     &Patchcore$_{BERT}$ (CVPR2022) & 0.461      & 0.621                          & 0.541          & 0.550         & 0.529 & 0.538 & 0.500   & 0.533 & 0.481 & 0.581 & 0.497 & 0.524&\multicolumn{2}{|c}{0.567}\\
     &Reg3DAD (NIPS2023)  & 0.388      & 0.470                           & 0.396          & 0.629        & 0.519 & 0.576 & 0.482 & 0.564 & 0.519 & 0.610  & 0.503 & 0.455&\multicolumn{2}{|c}{0.527}\\
     &IMR (CVPR2024)  &0.603      & \underline{0.676}                          & \underline{0.696}          & 0.533        & \underline{0.757} & 0.614 & 0.700 & 0.524 & 0.676 & 0.635 & 0.630  & \underline{0.594}&\multicolumn{2}{|c}{0.659}\\
     &R3DAD (ECCV2024)  & \underline{0.696}      & \textbf{0.736}                          & \textbf{0.900}          & \underline{0.788}        & 0.729 & \underline{0.752} & \underline{0.742} & 0.630 & \textbf{0.757} & \underline{0.771} & \underline{0.721} & \textbf{0.718}&\multicolumn{2}{|c}{\underline{0.749}}\\
     &\textbf{PLANE} & \textbf{0.759}      & 0.467                          & 0.652          & \textbf{0.896}        & \textbf{0.771} & \textbf{0.971} & \textbf{0.782} & \textbf{0.773} & \underline{0.690}  & \textbf{0.938} & \textbf{0.964} & 0.592 & \multicolumn{2}{|c}{\bf{0.836$\pm$0.006}}\\
    \bottomrule
  \end{tabular}}
  \caption{Object-level anomaly detection AUROC of 40 categories for the Anomaly-ShapeNet dataset. The best results are highlighted in \textbf{bold}, while the second-best results are \underline{underlined}.}
  \label{tab:1}
\end{table}

\begin{figure}[]
\centering
\includegraphics[width=0.7\columnwidth]{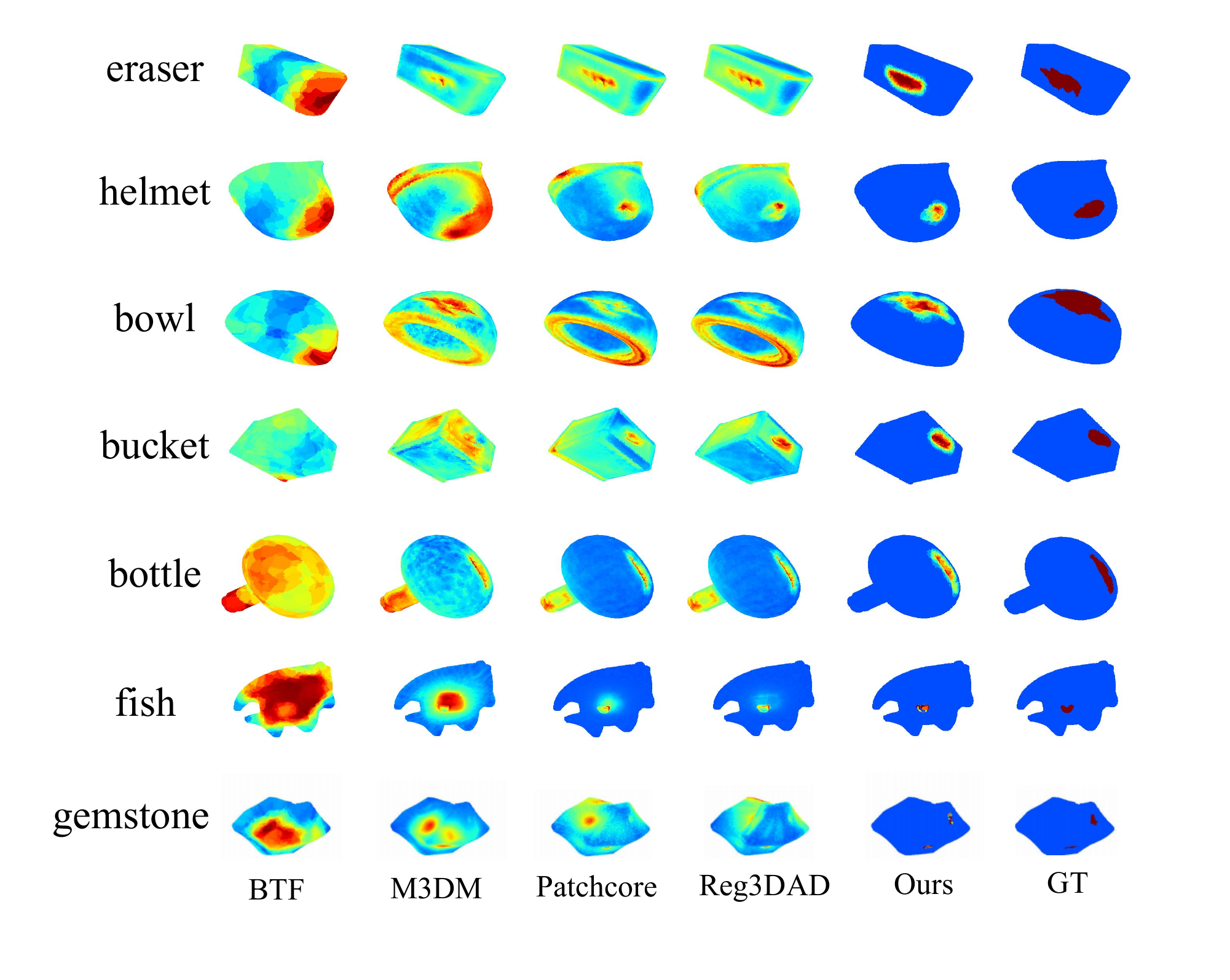}
\caption{Qualitative results visualization of anomaly localization performance on the Anomaly-ShapeNet dataset and the Real3D-AD datasets.}
\label{fig:4}
\end{figure}

\subsection{Implementation Details} We adopt the publicly available ULIP2-PointBert \cite{xue2024ulip} as the algorithm's backbone. All input point clouds are randomly downsampled to a fixed size of 2048 points. The length of learnable text prompts is set to 6. We select \{2, 5, 8, 11\} layers as intermediate outputs used in calculating the anomaly localization. The Adam optimizer is used, with a learning rate of 0.0001 for the prompt feature adaptation module, 0.00001 for the dual-prompts creator module and text learnable prompts. The model is trained with a batch size of 4 over 1400 epochs. All the experiments are performed on a single NVIDIA RTX 3090 GPU with 24GB. All results reported below are averaged over three independent runs with different random seeds.

\subsection{Comparison with SOTAs} The comparative methods employed in this study are as follows: four embedding-based methods, namely BTF \cite{horwitz2023BTF}, M3DM \cite{wang2023m3dm}, Patchcore \cite{roth2022patchcore}, and Reg3DAD \cite{liu2024real3d}, as well as two reconstruction methods, namely
IMR \cite{IMR} and R3DAD \cite{r3dad}\footnote[1]{Since there is no official implementation of the IMR method and the R3DAD method, we report the results from their original papers.}. Here, the M3DM$_{MAE}$ uses PointMAE \cite{pang2022pointmae} as the point cloud feature extractor, whereas M3DM$_{Bert}$ instead uses PointBert \cite{yu2022pointbert} as the extractor. The representation method for Patchcore is also the same. It is worth noting that the results reported for the other methods in the table are based on one-class-one-model paradigm, whereas we use a single model to perform detection across all classes, which is not only more
prevalent but also more valuable in practical industrial scenarios.


\textbf{Qualitative Results.} We conduct substantial qualitative experiments on the Anomaly-Shapenet and the Real3D-AD datasets to visually demonstrate the superiority of the proposed method. Fig. \ref{fig:4} shows the anomaly localization results of the PLANE method as compared to that of the other approaches. From the figure, it is evident that methods using hand-crafted operators or pre-trained models to extract features (such as BTF, M3DM, PatchCore) are unable to effectively localize anomalies in space, and that the predicted anomaly regions are often much larger than the actual ones. In contrast, the proposed our method could locate the anomalous regions accurately both for the Anomaly-ShapeNet and the Real3D-AD datasets.

\textbf{Quantitative Results on the Anomaly-ShapeNet.} The AD results for each category for the Anomaly-ShapeNet are shown in Table \ref{tab:1}. Across 40 categories, it is clear that the proposed PLANE exceeds the state-of-the-art by 8.7\% in Object-AUROC, reaching 0.836. The complete experimental results are displayed in Table \ref{tab:3},  which demonstrate that PLANE consistently achieved the best performance among the seven evaluation metrics. For point-level performance, PLANE surpass Reg3DAD by 33\% in PRO and 27.9\% in AP. The significant improvements across various metrics demonstrate that the proposed method has strong and stable AD and localization performance on this dataset.

\textbf{Quantitative Results on the Real3D-AD.} We also validate the method on the Real3D-AD dataset. Table \ref{tab:2} presents the object-level AUROC anomaly detection result by category, while the results for the seven metrics are shown in Table 3. Again, the PLANE demonstrates strong performance, for instance, in Object-AUROC, PLANE achieves 0.777, surpassing the second place R3DAD by 4.3\%. It can further be noted that in point-level metrics, PLANE also performs well, achieving 0.736, 0.511, and 0.201 in Point-AUROC, PRO, and AP, respectively, outperforming the other considered methods.

\begin{table}
  \centering
  \scriptsize
  \renewcommand\arraystretch{1.0}
  \resizebox{1\textwidth}{!}{
  \begin{tabular}{@{}cl|m{0.65cm}<{\centering}m{0.65cm}<{\centering}m{0.65cm}<{\centering}m{0.65cm}<{\centering}m{0.65cm}<{\centering}m{0.65cm}<{\centering}m{0.65cm}<{\centering}m{0.65cm}<{\centering}m{0.65cm}<{\centering}m{0.65cm}<{\centering}m{0.65cm}<{\centering}m{0.65cm}<{\centering}|m{1.95cm}<{\centering}}
    \toprule
    & Method & Airplane & Car & Candy & Chicken & Diamond & Duck & Fish & Gemstone & Seahorse & Shell & Starfish& Toffees & Mean\\
    \midrule 
     &BTF (CVPR2023)    & 0.520        & 0.560                           & 0.462          & 0.405        & 0.545       & 0.784       & 0.549       & 0.648       & \underline{0.779}       & 0.754       & 0.575       & 0.630&0.601 \\
     &M3DM$_{MAE}$ (CVPR2023)   & 0.438       & 0.572                          & 0.486          & 0.537        & 0.608       & 0.460        & 0.492       & 0.614       & 0.509       & 0.658       & 0.543       & 0.565       & 0.540 \\
     &M3DM$_{BERT}$ (CVPR2023) & 0.381       & 0.490                           & 0.466          & 0.540         & 0.621       & 0.488       & 0.520        & 0.592       & 0.498       & 0.538       & 0.494       & 0.522       & 0.513 \\
     &Patchcore$_{MAE}$ (CVPR2022) & 0.664       & 0.510                           & 0.546          & 0.650         & 0.771       & 0.579       & 0.696       & 0.396       & 0.572       & 0.499       & 0.495       &\underline{0.721}       & 0.592 \\
     &Patchcore$_{BERT}$ (CVPR2022) & 0.560        & 0.474                          & 0.489          & 0.579        & 0.676       & 0.538       & 0.682       & 0.476       & 0.542       & 0.427       & 0.508       & 0.618       & 0.547\\
     &Reg3DAD (NIPS2023)   & 0.697       & 0.706                          & \underline{0.760}          & \underline{0.719}        & \underline{0.921}        & 0.555       & \underline{0.928}       & 0.434       & \textbf{0.827}       & 0.557         & 0.483       & 0.696       & 0.690\\
     &IMR (CVPR2024) & \underline{0.762}       & \underline{0.755}                          & 0.711          & \textbf{0.780}        & 0.905       & 0.517       & 0.880        & \underline{0.674}       & 0.604       & 0.665       & \underline{0.674}       & \textbf{0.774}       & 0.725 \\
     &R3DAD (ECCV2024)    & \textbf{0.772}       & 0.696                          & 0.713          & 0.714        & 0.685       & \textbf{0.909}       & 0.692       & 0.665       & 0.720       & \underline{0.840}       & \textbf{0.701}       & 0.703       & \underline{0.734}\\
      &\textbf{PLANE} & 0.630       & \textbf{0.774}                         & \textbf{0.833}         & 0.693  & \textbf{0.998} & \underline{0.839} & \textbf{0.939} & \textbf{0.891} & 0.540 & \textbf{0.866} & 0.601 & 0.716&\bf{0.777$\pm$0.005}\\
    \bottomrule
  \end{tabular}}
  \caption{Object-level anomaly detection AUROC of 12 categories for the Real3D-AD dataset. The best results are highlighted in \textbf{bold}, while the second-best results are \underline{underlined}.}
  \vspace{3pt}
  \label{tab:2}
\end{table}

\begin{table}[]
\centering
\resizebox{1\textwidth}{!}{
\renewcommand{\arraystretch}{1.2} 
\begin{tabular}{l|ccccccc|ccccccc}
\hline\noalign{\vskip 0.5mm} 
                    & \multicolumn{7}{c|}{Anomaly-ShapeNet}                           & \multicolumn{7}{c}{Real3D-AD}                          \\ 
\noalign{\vskip 0.5mm}\hline\noalign{\vskip 0.5mm}
Method              & O-AUROC & P-AUROC & P-PRO & O-AP & P-AP & O-F1 & P-F1 & O-AUROC & P-AUROC & P-PRO & O-AP & P-AP & O-F1 & P-F1 \\
BTF (CVPR2023)                   & 0.510    & 0.510    & 0.153 & 0.583 & 0.018 & 0.722 & 0.045 & 0.601     & 0.571   & 0.186 & 0.610  & 0.022 & 0.700   & 0.053 \\
M3DM$_{MAE}$ (CVPR2023)             & 0.551   & 0.651   & 0.266 & 0.620  & 0.029 & 0.731 & 0.057 & 0.540    & 0.631    & 0.276 & 0.547 & 0.023 & 0.684 & 0.050  \\
M3DM$_{BERT}$ (CVPR2023)             & 0.555   & 0.633   & 0.255 & 0.620 & 0.026  & 0.717 & 0.033 & 0.513   & 0.628   & 0.271 & 0.554 & 0.025 & 0.669 & 0.044 \\
Patchcore$_{MAE}$(CVPR2022)          & 0.547   & 0.645   & 0.221 & 0.618 & 0.028 & 0.725 & 0.057 & 0.592   & 0.643   & 0.297 & 0.602 & 0.057 & 0.687 & 0.104 \\
Patchcore$_{BERT}$ (CVPR2022)       & 0.567   & 0.648   & 0.221 & 0.633 & 0.033 & 0.713 & 0.062 & 0.547   & 0.614   & 0.267 & 0.564 & 0.038 & 0.679 & 0.081 \\
Reg3DAD (NIPS2023)                    & 0.527   & 0.632   & 0.219 & 0.604 & 0.024 & 0.720  & 0.069 & 0.690   & 0.695   & 0.306 & 0.709 & 0.115 & 0.731 & 0.164 \\
IMR (CVPR2024)                     & 0.659   & 0.650    & -     & -     & -     & -     & -     & 0.725   & -       & -     & -     & -     & -     & -     \\
R3DAD (ECCV2024)                    & 0.749   & -       & -     & -     & -     & -     & -     & 0.734   & -       & -     & -     & -     & -     & -     \\
\textbf{PLANE}    & \textbf{0.836$\pm$0.006} & \textbf{0.821$\pm$0.004} & \textbf{0.549$\pm$0.003} & \textbf{0.875$\pm$0.005} & \textbf{0.303$\pm$0.003} & \textbf{0.850$\pm$0.004} & \textbf{0.348$\pm$0.002}  & \textbf{0.777$\pm$0.005} & \textbf{0.736$\pm$0.004}   & \textbf{0.511$\pm$0.006} & \textbf{0.810$\pm$0.005} & \textbf{0.201$\pm$0.004} & \textbf{0.770$\pm$0.003} & \textbf{0.254$\pm$0.004} \\
\noalign{\vskip 0.5mm}\hline 
\end{tabular}}
\caption{ More comprehensive results on the Anomaly-ShapeNet and the Real3D-AD datasets. The best results are highlighted in \textbf{bold}.}
\label{tab:3}
\end{table}

\begin{figure}[!t]
\centering
\includegraphics[width=0.7\columnwidth]{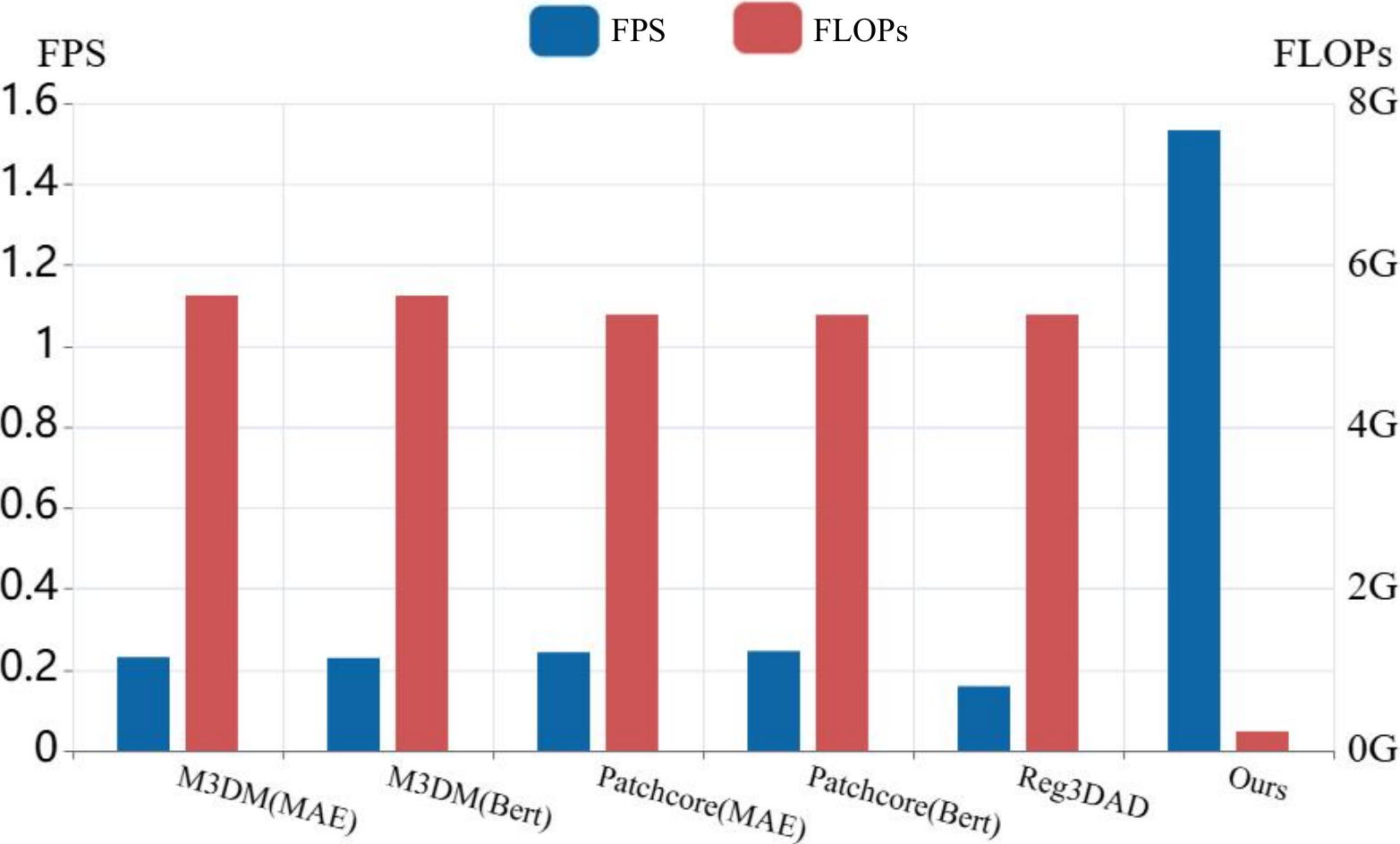}
\caption{The inference speed (FPS) and computational overhead (FLOPs) of various 3D AD methods are compared on the Real3D-AD dataset. A higher FPS indicates better performance, while a lower FLOPs value is preferable.}
\label{fig:5}
\end{figure}

\textbf{Computational Efficiency.} We compared the computational complexity and inference efficiency of the PLANE model with other existing 3D detection methods. We use floating-point operations (FLOPs) as the computation metric and frames per second (FPS) as the inference efficiency metric. As shown in Fig. \ref{fig:5}, the proposed method has significantly lower computational complexity as compared to the other methods. Moreover, in terms of inference time, the proposed method achieves an FPS of 1.53, clearly outperforming the other methods. This improvement is due to the PLANE model's use of a highly generalized point cloud text model, which eliminates the need for a memory bank, and significantly reducing the computational overhead required for matching in a memory bank. As a result, the proposed method not only demonstrates strong detection capabilities but also achieves lower computational complexity and higher inference speed.

\subsection{Ablation Study}

To investigate the effectiveness of individual components, we conduct ablation studies using the Anomaly-ShapeNet dataset. 

\textbf{Effects of the Key Components.}
We evaluate the effectiveness of our components, including the pseudo-anomaly sample generator (Ano3D), the point cloud feature adaptation (PCFA), and the dynamic prompts creator module (DPCM). As shown in Table \ref{tab:4}, the experimental results indicate that the pseudo-anomaly generation addresses the issue of limited normal samples, significantly improving the performance. Furthermore, PCFA can be seen to bring a large improvement of 8.8\%$\uparrow$ on the Object-AUROC. The DPCM also leads to performance enhancements, boosting the Object-AUROC by 2.3\% to reach 0.836, and improving the Point-AUROC by 1.1\%, bringing it to 0.821.
\begin{table}[h]
\centering
\resizebox{0.6\textwidth}{!}{
\begin{tabular}{ccc|cc}
\hline
\ Ano3D & PCFA & \multicolumn{1}{l|}{DPCM} & Object-AUROC & {Point-AUROC} \\ \hline
    $\times$                           &     $\times$              &      $\times$              & 0.517        & 0.533                          \\
  $\checkmark$                              &       $\times$            &        $\times$           & 0.725        & 0.809                            \\
   $\checkmark$                             &   $\checkmark$              &         $\times$          & 0.813        & 0.810                            \\
     $\checkmark$                           &       $\checkmark$          &       $\checkmark$          & 0.836        & 0.821                            \\ \hline
\end{tabular}}
\caption{Ablation studies on key components.}
\label{tab:4}
\end{table}

\begin{figure}[h]
\centering
\includegraphics[width=0.7\columnwidth]{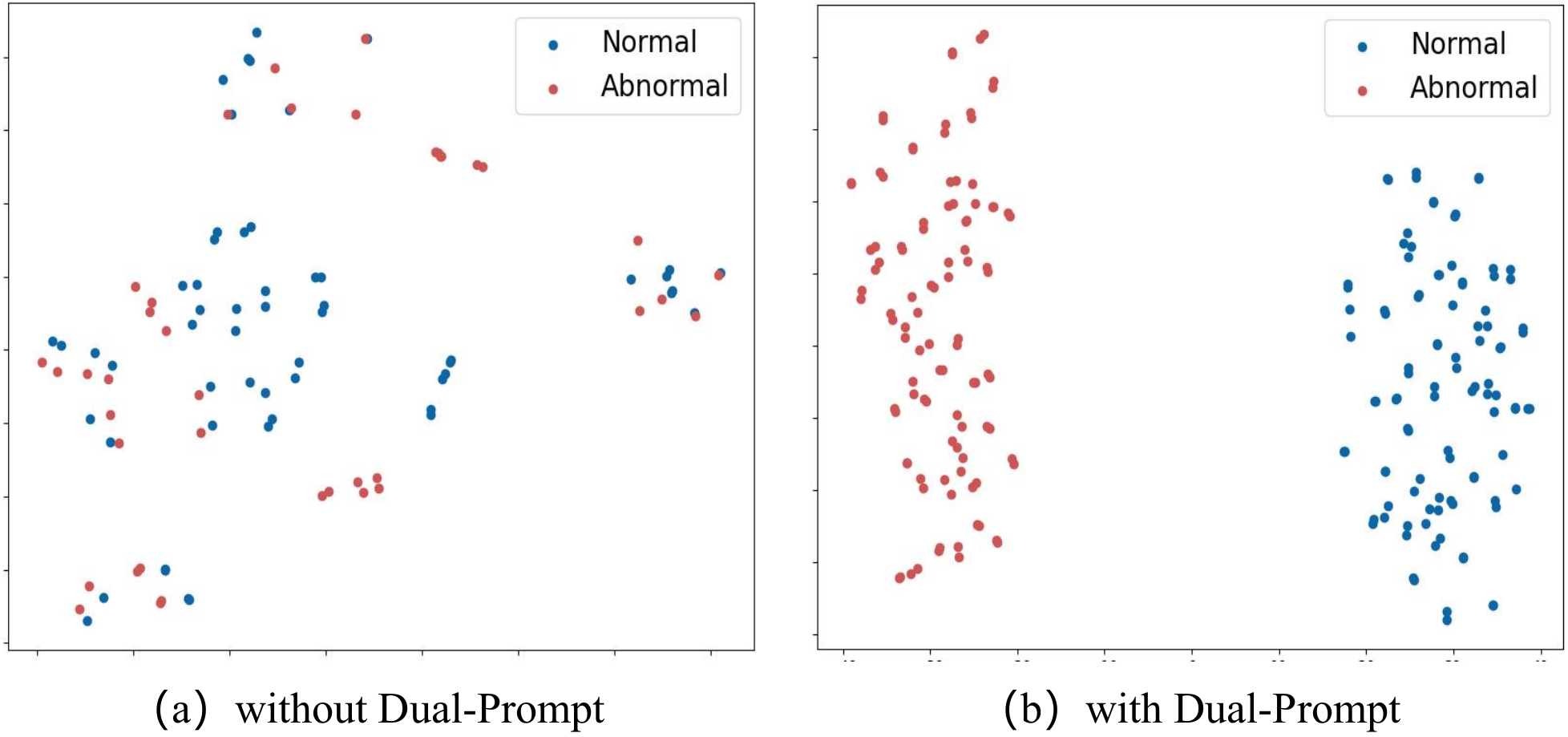}
\caption{Visualization of normal/ anomalous prompt features using t-SNE. (a) The original text prompt in feature space. (b) The text prompt (using dual-propmts) in feature space.}
\label{fig:6}
\end{figure}

\textbf{Effect of the Text Prompt.}
We use t-SNE to visualize the distances between normal and anomalous prompt embeddings within the feature space. Fig. \ref{fig:6}(a) illustrates that there are no clear boundaries separating different types of text prompts, which significantly impairs pre-trained model  performance in the downstream AD tasks. In contrast, Fig. \ref{fig:6}(b) shows that with text prompt tunning, distinct boundaries emerge between different types of text prompts in the feature space, effectively illustrating its benefits.

\textbf{Effect of Point Cloud Prompt.}
We further compare the proposed point cloud prompt learning with existing visual prompt tunning (VPT) methods \cite{jia2022visual}. The experimental results are summarized in Table \ref{tab:6}, where VPT-shallow implies that visual prompts are added only at the first layer whereas in VPT-Deep, the visual prompts are added at each layer. The results demonstrate that the proposed prompt learning method has better performance for both anomaly detection and localization.

\begin{table}[h]
\tiny
\centering
\resizebox{0.6\textwidth}{!}{
\begin{tabular}{c|cc}
\hline
\multicolumn{1}{l|}{Methods} & Object-AUROC  & Point-AUROC \\ \hline
                    VPT-shallow       & 0.811          & 0.810     \\
                    VPT-deep      & 0.824         & 0.804      \\
                    Ours        & 0.836       & 0.821       \\ \hline
\end{tabular}}
\caption{Comparison with existing vision prompt tunning methods.}
\label{tab:6}
\end{table}

\begin{figure}[!t]
\centering
\includegraphics[width=0.75\columnwidth]{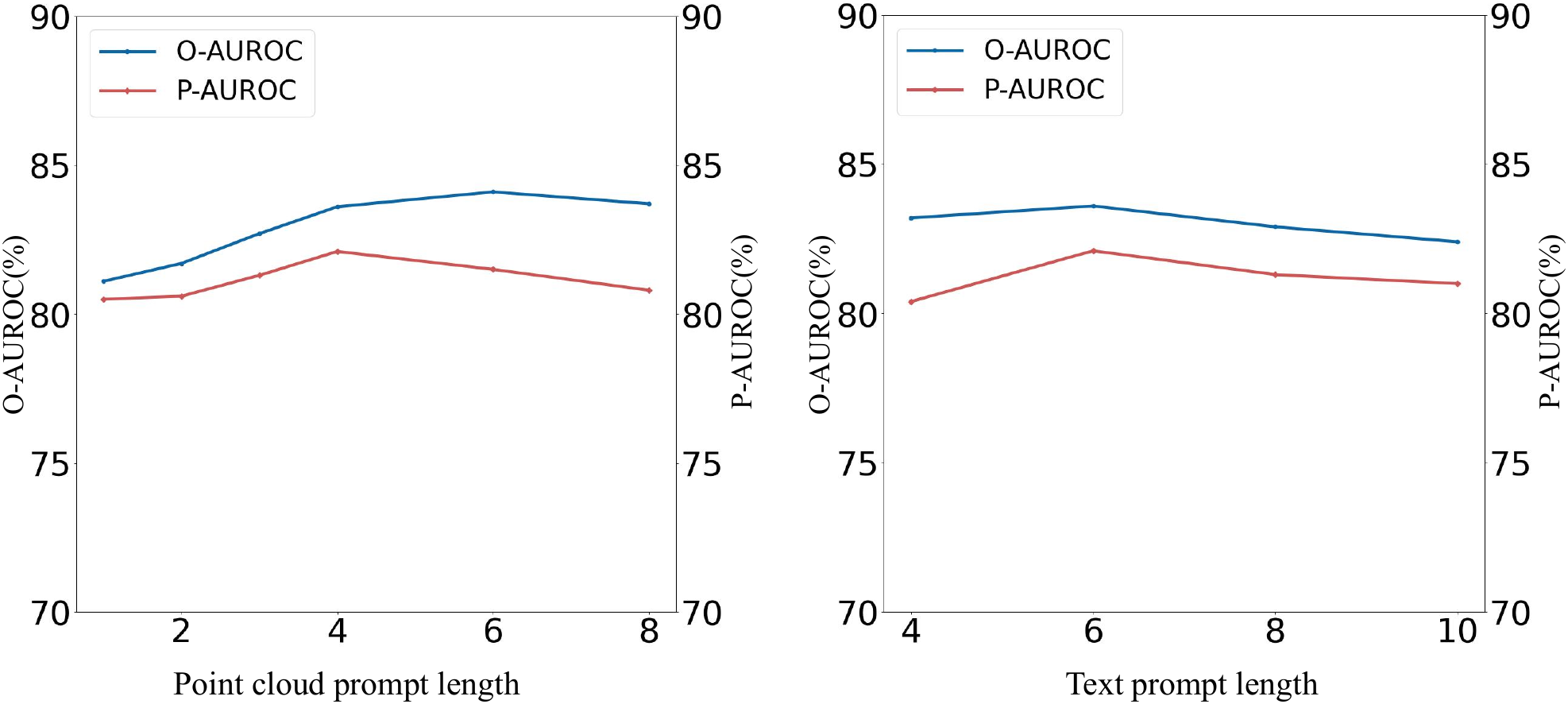}
\caption{Ablation studies on different prompt length for point cloud prompts and text prompts.}
\label{fig:8}
\end{figure}

\textbf{Study on the static prompt length.} We investigate the impact of the length of the point cloud and text prompts on model performance. As shown in Fig. \ref{fig:8}, when the value of length is relatively small, the detection and localization results improve as the length increases. However, as the length of the prompts continues to increase, there is a decline in performance, suggesting that overly complex prompts may contain redundant information. Therefore, selecting an optimal length, such as N = 6 for text prompts and M = 4 for point cloud prompts, helps the model learn effective prompts.

\textbf{Effects of the Focal Loss and Dice Loss.} 
We also conduct an ablation study on the loss functions. The experimental results indicate that both loss functions are effective for PLANE, as they mitigate the imbalance between anomalous and normal points caused by the small size of the anomalous regions. Table \ref{tab:5} shows the quantitative results. The method achieves 0.744 Object-AUROC and 0.593 Point-AUROC using only the dice loss, while achieving 0.625/0.769 with only the focal loss. Using both loss functions simultaneously yields the best experimental results, at 0.836 Object-AUROC and 0.821 Point-AUROC.

\begin{table}[h]
\centering
\resizebox{0.6\textwidth}{!}{
\begin{tabular}{cc|cc}
\hline
Focal loss & \multicolumn{1}{l|}{Dice loss} & Object-AUROC  & Point-AUROC  \\ \hline
  $\checkmark$        &                                & 0.625          & 0.769        \\
           &        $\checkmark$                     & 0.744          & 0.593        \\
  \checkmark         &   \checkmark                             & 0.836          & 0.821        \\ \hline
\end{tabular}}
\caption{Ablation study on the loss function.}
\label{tab:5}
\end{table}

\textbf{Effect of multi-layer features.} 
The effectiveness of multi-layer feature aggregation is studied, and the results are presented in Table \ref{tab:6}. 
The results indicate that multi-layer features \{2, 5, 8, 11\} outperform single-layer \{2\} or double-layer \{2, 5\} features. It is suggested a single intermediate layer feature has limited information and is not enough to express the complete defect features, while the combination of local information from shallow features and global information from deep features is more conducive to the detection of anomalies. Therefore, multi-layer features enhance both anomaly detection and localization performance.

\begin{table}[h]
\tiny
\centering
\resizebox{0.6\textwidth}{!}{
\begin{tabular}{ccc}
\hline
\multicolumn{1}{l|}{Layers} & Object-AUROC & Point-AUROC \\ \hline
                    \{2\}        & 0.769        & 0.766     \\
                    \{2,5\}       & 0.803       & 0.807      \\
                    \{2,5,8\}        & 0.831        & 0.810      \\
                    \{2,5,8,11\}       & 0.836      & 0.821    \\ \hline
\end{tabular}}
\caption{Ablation study on multi-layer feature aggregation.}
\label{tab:6}
\end{table}

\section{Conclusion}
In this paper, we propose a 3D point cloud AD method based on pre-trained PLMs called PLANE, which detects anomalies by measuring the similarity between point cloud and text features. We propose a dynamic-static dual-prompts approach that enables the PLMs to transfer knowledge to the downstream AD tasks. Additionally, a pseudo-anomaly generation method based on point cloud data characteristics is introduced for data augmentation, thereby enhancing the model's generalization capabilities.
Experiments demonstrate that the proposed multi-class-one-model method outperforms existing methods following the one-class-one-model paradigm on the Anomaly-ShapeNet and the Real3D-AD datasets.








\bibliographystyle{elsarticle-num} 
\bibliography{reference}

\end{document}